\documentclass[10pt,twocolumn,letterpaper]{article}

\usepackage[pagenumbers]{iccv} 

\makeatletter
\DeclareRobustCommand\onedot{\futurelet\@let@token\@onedot}
\def\@onedot{\ifx\@let@token.\else.\null\fi\xspace}
\def\eg{e.g\onedot} 
\def\ie{i.e\onedot}

\makeatother

\newcommand{\boldparagraph}[1]{\vspace{0.2cm}\noindent{\bf #1:}}

\definecolor{darkgreen}{rgb}{0,0.7,0}

\usepackage{graphicx}
\usepackage{amsmath}
\usepackage{amssymb}
\usepackage{booktabs}
\usepackage{pifont}

\usepackage{algorithm}
\usepackage{algpseudocode}
\algnewcommand\algorithmicinput{\textbf{Input:}}
\algnewcommand\Input{\item[\algorithmicinput]}

\usepackage{times}
\usepackage{epsfig}
\usepackage{graphicx}
\usepackage{amssymb}

\usepackage{float}
\usepackage{booktabs}
\usepackage{caption}
\usepackage{subcaption}
\usepackage{multirow}
\usepackage[dvipsnames]{xcolor}

\mathchardef\mhyphen="2D
\usepackage[utf8]{inputenc}
\usepackage{comment}
\usepackage{tabularx,colortbl}
\usepackage{xparse,mathtools}
\usepackage{diagbox}
\usepackage{adjustbox,lipsum}

\usepackage{breqn}

\newcommand{\cmark}{\textcolor{my_green}{\ding{51}}} 
\newcommand{\xmark}{\textcolor{my_red}{\ding{55}}} 

\usepackage{xcolor}         
\usepackage{color, colortbl}
\definecolor{citecolor}{HTML}{2980b9}
\definecolor{linkcolor}{HTML}{c0392b}
\definecolor{darkorange}{HTML}{FF8C00}
\definecolor{chocolate}{HTML}{D2691E}
\definecolor{darkgreen}{HTML}{006400}
\definecolor{darkblue}{HTML}{00008B}
\definecolor{mediumblue}{HTML}{0000CD}
\definecolor{dodgerblue}{HTML}{1E90FF}
\definecolor{royalblue}{HTML}{4169E1}
\definecolor{shadecolor}{RGB}{237,237,237}
\definecolor{backred}{RGB}{255, 190, 190}
\definecolor{backblue}{RGB}{210, 230, 250}

\definecolor{zrrgreen}{HTML}{008000}
\definecolor{zrrblue}{HTML}{4682B4}
\definecolor{zrrred}{HTML}{B22222}

\definecolor{my_green}{RGB}{51,102,0}
\definecolor{my_red}{RGB}{204, 0, 0}

\usepackage{tabu}
\usepackage{tabulary,overpic}

\newlength\savewidth

\definecolor{cvprblue}{rgb}{0.21,0.49,0.74}
\usepackage[pagebackref,breaklinks,colorlinks,allcolors=cvprblue]{hyperref}

\colorlet{lr}{red!70!white}
\colorlet{lb}{cyan!70!white}
\colorlet{lo}{orange!70!white}
\colorlet{lm}{magenta!70!white}

\definecolor{lb}{HTML}{FFEFCC}
\definecolor{lo}{HTML}{CCFFEF}
\definecolor{lm}{HTML}{CDCCFF}

\definecolor{g0}{HTML}{DEFFCE}
\definecolor{g1}{HTML}{98FB98}

\definecolor{v0}{HTML}{D3E0FF}
\definecolor{v1}{HTML}{89CFF0}

\usepackage[capitalize]{cleveref}
\crefname{section}{Sec.}{Secs.}
\Crefname{section}{Section}{Sections}
\Crefname{table}{Table}{Tables}
\crefname{table}{Tab.}{Tabs.}

\def\paperID{3205} 
\def\confName{ICCV}
\def\confYear{2025}

\title{Passing the Driving Knowledge Test}

\author{
  Maolin Wei$^{1}$\thanks{Equally contributed.} \quad 
  Wanzhou Liu$^{2}$\footnotemark[1] \quad 
  Eshed Ohn-Bar$^{1}$ \\
  $^{1}$Boston University \quad
  $^{2}$Washington University in St. Louis
}

\begin{document}
\maketitle

\begin{abstract}

If a Large Language Model (LLM) were to take a driving knowledge test today, would it pass? Beyond standard spatial and visual question-answering (QA) tasks on current autonomous driving benchmarks, driving knowledge tests require a complete understanding of all traffic rules, signage, and right-of-way principles. To pass this test, human drivers must discern various edge cases that rarely appear in real-world datasets. 
In this work, we present \textbf{DriveQA}, an extensive open-source text and vision-based benchmark that exhaustively covers traffic regulations and scenarios. 
Through our experiments using DriveQA, we show that (1) state-of-the-art LLMs and Multimodal LLMs (MLLMs) perform well on basic traffic rules but exhibit significant weaknesses in numerical reasoning and complex right-of-way scenarios, traffic sign variations, and spatial layouts, (2) fine-tuning on DriveQA improves accuracy across multiple categories, particularly in regulatory sign recognition and intersection decision-making, (3) controlled variations in DriveQA-V provide insights into model sensitivity to environmental factors such as lighting, perspective, distance, and weather conditions, and (4) pretraining on DriveQA enhances downstream driving task performance, leading to improved results on real-world datasets such as nuScenes and BDD, while also demonstrating that models can internalize text and synthetic traffic knowledge to generalize effectively across downstream QA tasks. Project page: \href{https://driveqaiccv.github.io}{https://driveqaiccv.github.io}.
\end{abstract}

\begin{figure}[t!]
    \centering
      \includegraphics[trim={0cm 2.3cm 5.8cm 0cm},clip,width=\linewidth]{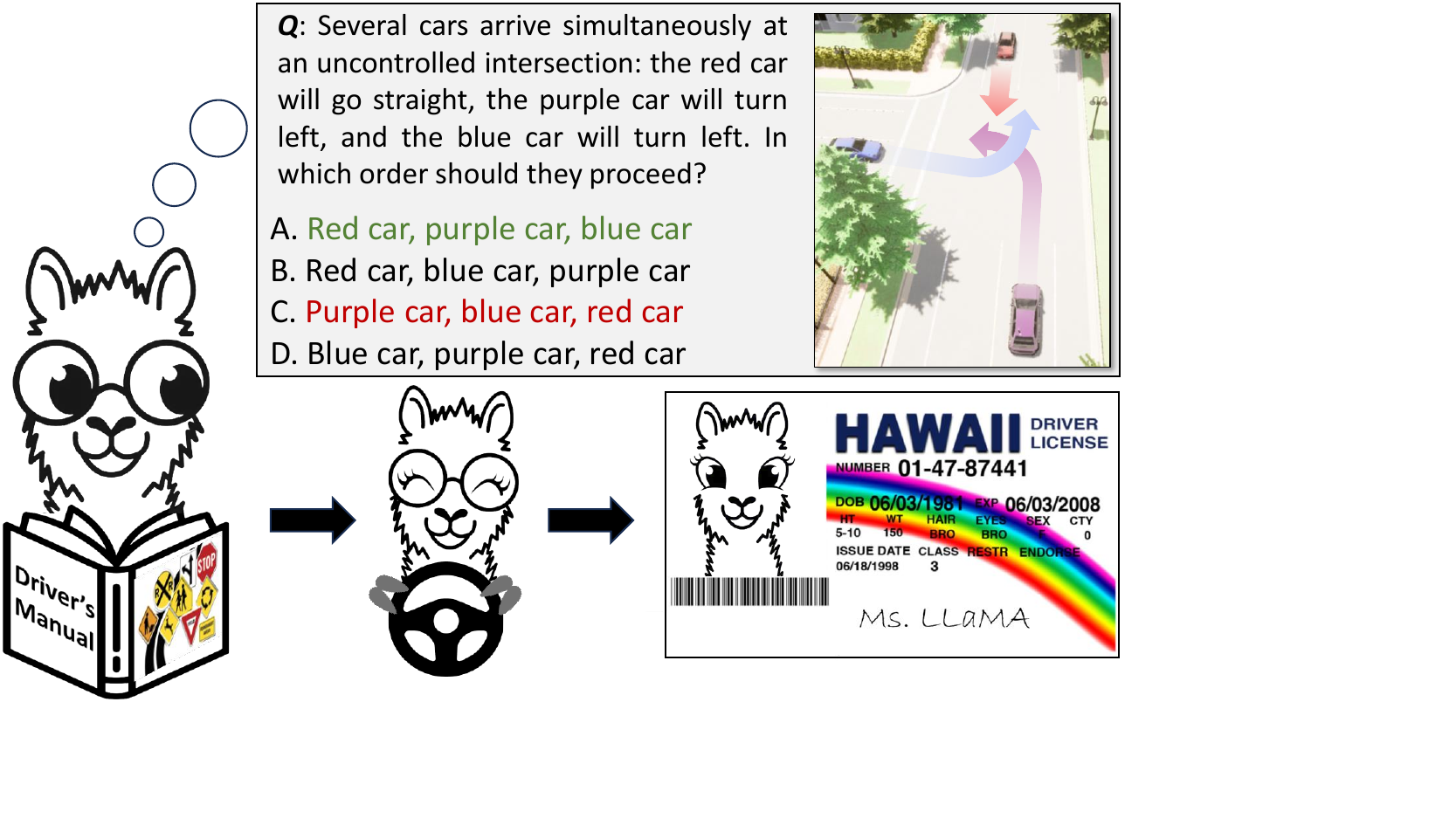} 
    \vspace{-20pt}
    \caption{\textbf{Can LLMs Pass a Driving Knowledge Test?} We introduce a comprehensive multimodal dataset to evaluate the traffic rule-following capabilities of MLLMs. While most question answering (QA) benchmarks in autonomous driving focus on spatial understanding and common planning tasks, our \textbf{DriveQA} dataset assesses broad driving knowledge. The challenging benchmark comprises text-based questions derived from various U.S. state driving manuals, as well as visual tasks for traffic sign recognition and right-of-way judgment. We evaluate both text-only and image-text QA using synthetic images (varying perspectives, weather, time of day, and sign types) while showing transferability and generalization to downstream real-world driving tasks. Ths figure shows one example for a right-of-way question, a category where models frequently struggle. Incorrect responses are highlighted in {\color{red} red} and correct answers in {\color{darkgreen} green}.
    }
    \label{fig:intro}
    \vspace{-0.4cm}
\end{figure}

\begin{table*}[t!]
\centering
\caption{\textbf{A Multimodal Dataset Emphasizing  Traffic Rules.} The table compares existing benchmarks in terms of: the total number of images (\textbf{\#Images}), the number of QA pairs (\textbf{\#QA Pairs}), the method of annotation (\textbf{Annotations}, A+M means semi-automatic labeling), environment settings (\textbf{Settings}, including camera perspective of forward-\textbf{Fwd}, oblique-\textbf{Obl}, and top-down-\textbf{Top} views, weather, and time of day conditions), explanations for each question's answer (\textbf{Explanations}), and traffic rule reasoning (\textbf{Traffic Rules}), which is our focus.
}
\label{tab:comparison}
\vspace{-5pt}
\begin{adjustbox}{max width=\textwidth}
\begin{tabular}{l | c c c c c c c c c c}
\toprule
\multirow{2}{*}{\textbf{Benchmarks}} & 
\multirow{2}{*}{\textbf{Image Source}} & 
\multirow{2}{*}{\textbf{\#Images}} & 
\multirow{2}{*}{\textbf{\#QA Pairs}} & 
\multirow{2}{*}{\textbf{Annotations}} & 
\multicolumn{3}{c}{\textbf{Settings}} &
\multirow{2}{*}{\textbf{Explanations}} & 
\multirow{2}{*}{\textbf{Traffic Rules}} 
\\
\cmidrule(lr){6-8}
& & & & & \textbf{Perspective} & \textbf{Weather} & \textbf{Time of Day} & & \\
\midrule
EQA-v1~\cite{das2018embodied}  & House3D~\cite{wu2018building} & 767 & 5,281  & A & Fwd, Obl & \xmark & \xmark & \xmark & \xmark\\
OpenEQA~\cite{majumdar2024openeqa} & OpenEQA & 180 & 1,600 & M & Fwd, Obl & \xmark & \xmark & \xmark & \xmark\\
SpatialVLM~\cite{chen2024spatialvlm} & Internet & 10M & 2B & A & Fwd, Obl & \xmark & \xmark & \xmark & \xmark &\\
NuScenes-QA~\cite{qian2023nuscenes} & nuScenes~\cite{caesar2020nuscenes} & 34K & 450K & A & Fwd & \cmark & \cmark & \xmark & \xmark &\\
DriveLM-nuScenes~\cite{sima2024drivelm} & nuScenes~\cite{caesar2020nuscenes} & 4,871 & 443K & A+M & Fwd & \cmark & \cmark & \xmark & \xmark\\
DriveLM-CARLA~\cite{sima2024drivelm} & CARLA~\cite{Dosovitskiy2017CORL} & 64,285 & 1,566K & A & Fwd & \cmark & \cmark & \xmark & \xmark\\
DriveBench~\cite{xie2025drivebench} & nuScenes~\cite{caesar2020nuscenes} & 19,200 & 20,498 & A+M & Fwd & \cmark & \cmark & \xmark & \xmark\\
LingoQA~\cite{marcu2023lingoqa} & LingoQA & 28k & 419.9K & A+M & Fwd & \cmark & \cmark & \xmark & \xmark\\
\midrule
\textbf{DriveQA-V (ours)} & CARLA~\cite{Dosovitskiy2017CORL}, Mapillary~\cite{neuhold2017mapillary}  & 68K & 448K & A+M & Fwd, Obl, Top & \cmark & \cmark & \cmark & \cmark \\
\textbf{DriveQA-T (ours)} & - & - & 26K & A+M & - & \cmark & \cmark & \cmark & \cmark\\ 
\bottomrule
\end{tabular}
\end{adjustbox}
\vspace{-5pt}
\end{table*}

\vspace{-0.2cm}
\section{Introduction}
\label{sec:intro}
Safe navigation in traffic requires not only recognizing and interpreting visual information but also reasoning over traffic rules and making decisions that align with regulations. To ensure drivers develop these critical skills, before receiving their license they must first pass a written knowledge test---a structured (multiple choice questions) assessment designed to evaluate precise understanding of traffic laws, right-of-way rules, and complex driving scenarios~\cite{cadmv,drivertests}. 

Driving tests are not merely procedural; they assess a driver's ability to apply reasoning across a wide range of traffic conditions. While primarily textual, these tests may also include graphical illustrations to ground questions in real-world scenarios. Recent advances in Multimodal Large Language Models (MLLMs)~\cite{dubey2024llama, team2024gemma, zhou2024tinyllava, lin2024vila, abdin2024phi} as general-purpose reasoning models provide an opportunity to explore a key question: how well do current vision-and-language models perform when faced with the same driving knowledge assessments? Even without targeted fine-tuning, MLLMs may inherit some traffic rule knowledge from their pretraining data (however, our findings indicate that both such knowledge and associated reasoning capabilities remain limited).  

Researchers have been increasingly integrating MLLMs into autonomous driving systems~\cite{mao2024a, mao2023gpt, xu2024drivegpt4, fu2024drive, wu2023language, cui2023drive, sha2023languagempc, chen2023driving,zhou2024embodied,tian2024tokenize,li2024driving,ma2024dolphins,zhang2024feedback,Lai2024UncertaintyGuidedNL}. However, while these models are often tested on perception-focused benchmarks that emphasize spatial awareness and standard planning tasks (\eg, lane keeping, collision avoidance~\cite{li2024ego,xing2025openemma,tian2024drivevlm}), they are rarely evaluated for their ability to understand and comply with diverse traffic regulations, such as reasoning about traffic rules, reacting safely to no-entry signs, or maintaining speed limit. While most existing datasets narrowly focus on perception and basic trajectory planning, driving knowledge tests are designed to assess a broad spectrum of all regulations, including rare traffic signs, difficult right-of-way cases, and edge-case rules that are essential for safe navigation but seldom appear in collected driving data. This highlights a critical gap in evaluating AI systems: while they may perform well in current benchmarks, their ability to reason over long-tail traffic rules and regulatory compliance remains understudied. There is also substantial anecdotal evidence suggesting that current commercial systems, e.g., Tesla's Full Self-Driving~\cite{x1,x2,x3,x4}, often struggle with interpreting traffic rules.
 
To address this gap and enhance the evaluation of reasoning capabilities in both LLMs and MLLMs, we introduce a novel driving knowledge benchmark, \textbf{DriveQA}. Our dataset includes both text-only question-answers (QA) and aligned image-text (VQA) pairs. Thus, we enable the first thorough evaluation of vision-and-language model performance across broad driving tasks, from basic regulatory questions and signs to complex multimodal reasoning tasks. Our \textbf{contributions} are summarized as follows:

\begin{itemize}
    \item  We introduce \textbf{DriveQA}, a large-scale benchmark featuring both text-based (\textbf{DriveQA-T}) and vision-based (\textbf{DriveQA-V}) driving knowledge assessments. To ensure broad coverage of traffic regulations, right-of-way rules, and rare driving scenarios, we leverage synthetic procedural data generation with comprehensive traffic reasoning, controlled variations (\eg, sign placement and weather), and new 3D sign assets integrated into CARLA~\cite{Dosovitskiy17CARLA}, as well as manually annotated real-world data from Mapillary~\cite{neuhold2017mapillary}.  DriveQA covers 19 question categories, 220 traffic signs, and 474K samples. 

\item We \textit{benchmark} state-of-the-art LLMs and MLLMs on DriveQA to uncover that while these models perform well on basic traffic rules, they struggle with numerical precision, right-of-way reasoning, spatial awareness, and environmental sensitivity (e.g., time-of-day, perspective, and geometric layouts). Our findings suggest that MLLMs inherit limited traffic knowledge from pretraining and require fine-tuning for our task. 

\item We demonstrate the effectiveness of DriveQA \textit{pretraining}; models trained on our text and purely synthetic data demonstrate improved performance across various real-world driving tasks~\cite{xing2025openemma, xu2020explainable}. We show that pretraining on DriveQA improves the performance on both trajectory prediction and driving action reasoning tasks. This highlights its role in evaluating and enhancing multimodal reasoning, and as a step toward bridging theory and practice in embodied AI systems that can learn to make decisions in the real-world based on text or synthetic data.

\end{itemize}

\begin{figure*}[t!]
    \centering
    \includegraphics[trim={1.72cm 5.8cm 1.6cm 5.8cm},clip,width=\linewidth]{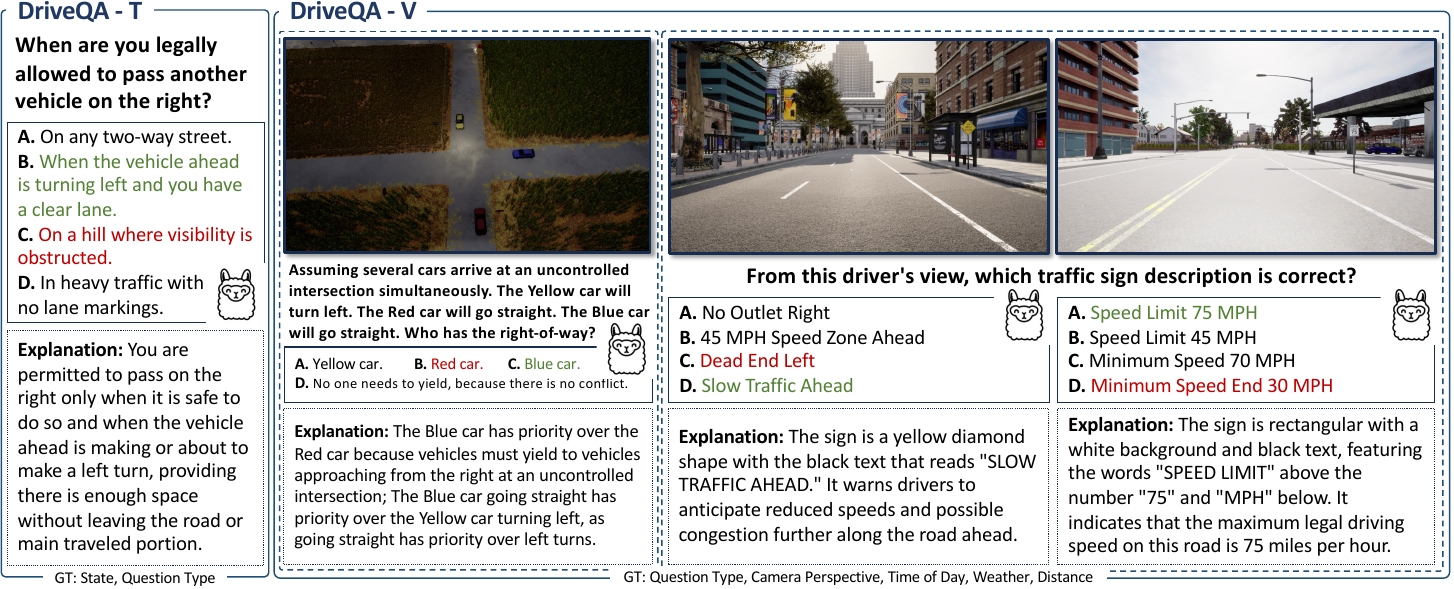}
    \vspace{-7mm}
    \caption{{\textbf{Example Questions and Answers of DriveQA Dataset.} We introduce a text and vision-based benchmark for extensively validating driving knowledge with question type categorization, answer explanation, and environmental information ground truth (GT). 
    }}
    \label{fig:DQA_data_examples}
    \vspace{-4mm}
\end{figure*}

\section{Related Work}
\label{sec:related}

Based on our survey of MLLM-based studies and VQA benchmarks for autonomous driving below, we find prior work rarely addressed traffic rules, signage, and right-of-way principles within their driving knowledge assessments. Relevant related benchmarks are compared in Table~\ref{tab:comparison}.

\boldparagraph{Multimodal Large Language Models}
Our study diagnoses multimodal reasoning capabilities in MLLMs~\cite{kim2024text,cui2024survey,peng2023kosmos,tong2024eyes,zhou2024tinyllava,tian2024drivevlm,wang2024omnidrive,zhang2024feedback}. A typical MLLM architecture comprises three main modules: a pre-trained modality encoder, a pre-trained language model, and a modality projector that aligns them. The modality encoder processes non-textual inputs, such as images, transforming them into representations compatible with the language models. Vision Transformer (ViT)~\cite{dosovitskiy2021an} is widely used to extract image features. For example, CLIP~\cite{radford2021learning} leverages ViT as its visual encoder to transform images into feature representations that align with text through extensive pre-training on large-scale image-text pairs. The modality projector aligns encoder outputs with the language model, enabling integration of modality data with text. A common approach is to use a set of learnable query tokens to extract information in a query-driven manner~\cite{carion2020end}, which has been employed by a variety of models~\cite{li2023blip,zhang2023video, dai2023instructblip, lai2025zerovo,chen2023x, yuan2021florence, chen2021pix2seq, zhai2022lit}. Additionally,  methods may design MLPs to transform the high-dimensional input features into a unified representation~\cite{liu2024visual, su2023pandagpt, pi2023detgpt,tong2024eyes}. Our systematic study controlling for variations in QA category and image factors reveals limitations of current alignment mechanisms in supporting multimodal or spatial reasoning.

\boldparagraph{MLLM-based Driving Agents}
While recent advancements have applied MLLMs to autonomous driving tasks, most focus on leveraging reasoning and language understanding capabilities to improve driving decisions in narrow tasks~\cite{chen2024driving, cui2024drive, cui2024receive, fu2024drive, mao2023gpt, sha2023languagempc, wen2023dilu, xu2024drivegpt4, zhou2024embodied}. For instance, several vision-and-language agents for motion planning and decision-making have been proposed and evaluated on datasets such as nuScenes~\cite{caesar2020nuscenes,mao2023gpt,xu2024drivegpt4,fu2024drive,achiam2023gpt,tian2024drivevlm,ma2024dolphins,mao2024a,li2024driving,renz2024carllava}. The key hypothesis in such studies is that MLLMs can inherit general-purpose reasoning and knowledge from pretraining; however, our findings suggest that while they may grasp basic traffic concepts, their ability to apply traffic reasoning in driving-specific scenarios remains limited. Moreover, these works have not explicitly addressed MLLMs' ability to comprehend diverse traffic rules and regulations-a critical requirement for safe driving. 

\boldparagraph{Datasets for Autonomous Driving} 
Several real-world, synthetic, and VQA benchmarks for autonomous driving are currently being used to evaluate driving models, including KITTI~\cite{geiger2012we,liao2022kitti}, Waymo Open~\cite{sun2020scalability}, Argoverse~\cite{chang2019argoverse,wilson2023argoverse}, and nuScenes~\cite{caesar2020nuscenes}. However, few incorporate more than a handful of traffic rules, \eg, researchers may evaluate collision on nuScenes~\cite{caesar2020nuscenes,zhang2023coaching,hu2023planning,tian2024drivevlm,dauner2023parting,Zhu2023LearningTD,zhang2022selfd}, yet lack coverage and exclude explicitly evaluating for traffic signs or right-of-way reasoning. Crowdsourced benchmarks such as Mapillary~\cite{neuhold2017mapillary}, which we augment with VQA annotations, are broad but still lack in long-tail events, motivating the use of synthetic benchmarks. Yet, prior simulation-based studies (\eg,~\cite{Dosovitskiy17CARLA,carlaleaderboard,jaeger2023hidden,Renz2022CORL,zhang2021end,shao2023reasonnet,fabbri2021motsynth,richter2017playing}) have only accounted for a handful of potential regulatory and safety violations. For instance, while CARLA~\cite{Dosovitskiy17CARLA} enables controllable and diverse data generation (\eg, perspectives, scenarios, weather), most traffic signs are missing in CARLA, a limitation addressed by our work. The development of MLLMs and their applications in autonomous driving lead to the emergence of driving vision-language datasets~\cite{qian2024nuscenes, marcu2023lingoqa, sima2024drivelm, arai2024covla, tian2024drivevlm, wang2024omnidrive, sachdeva2024rank2tell} specifically designed to support vision understanding and reasoning in complex driving scenarios. However, here as well existing efforts focus on scene understanding, perception and basic planning (\ie, collision avoidance, intersection boundary~\cite{li2024ego}), neglecting reasoning about traffic rules and regulations (\ie, reacting safely to no-entry signs, maintaining speed limit, etc.) which is a foundational driving test for humans. 

\begin{figure}[t!]
    \centering
    \includegraphics[trim={0.0cm 0 0.1cm 0},clip,width=\linewidth]{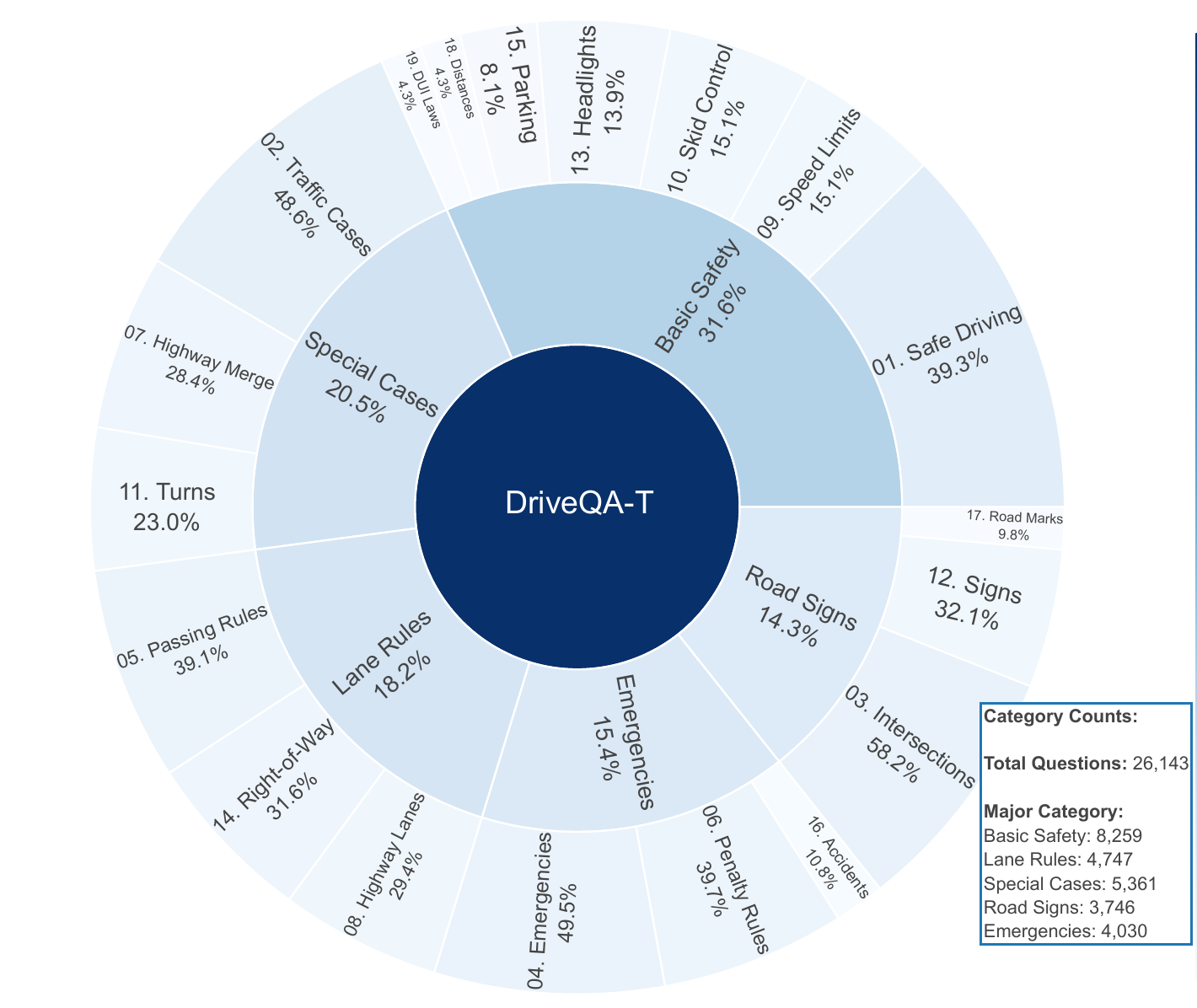}
    \vspace{-20pt}
    \caption{\textbf{Distribution of Question Type in DriveQA-T.} The benchmark covers five key domains and 19 sub-class types.}
    \label{fig:textdata_pie_chart}
    \vspace{-4mm}
\end{figure}

\vspace{-0.35cm}
\section{A Multimodal Driver Knowledge Test}
\label{sec:bench}
\vspace{-0.12cm}

In this section, we outline our scalable data collection and annotation process. Our dataset consists of QA pairs that cover essential aspects of real-world driving knowledge. As illustrated in Fig.~\ref{fig:DQA_data_examples}, our dataset comprises two tasks: DriveQA-T, which consists of text-based QA pairs on general driving rules, and DriveQA-V, focusing on visual (image-based) QA related to traffic sign comprehension and right-of-way scenarios. We adhere as closely as possible to standard driving knowledge tests to ensure meaningful comparisons to human performance on these assessments, and generate a diverse set of multiple-choice questions. We note that there are commercial driver knowledge tests available~\cite{drivertests,drivertests2}, however these are closed-source.
To ensure in-depth analysis, we further provide reasoning for ground-truth answers on both tasks. This design is intended to provide a holistic and systematic analysis of both LLMs and MLLMs in decision-oriented tasks. Ultimately, our overarching goal is to enable novel mechanisms to teach MLLMs real-world tasks, \eg, through text descriptions or synthetic examples. 

\begin{figure}[!t]
    \centering
    \includegraphics[trim={5cm 0 5cm 0},clip,width=\linewidth]{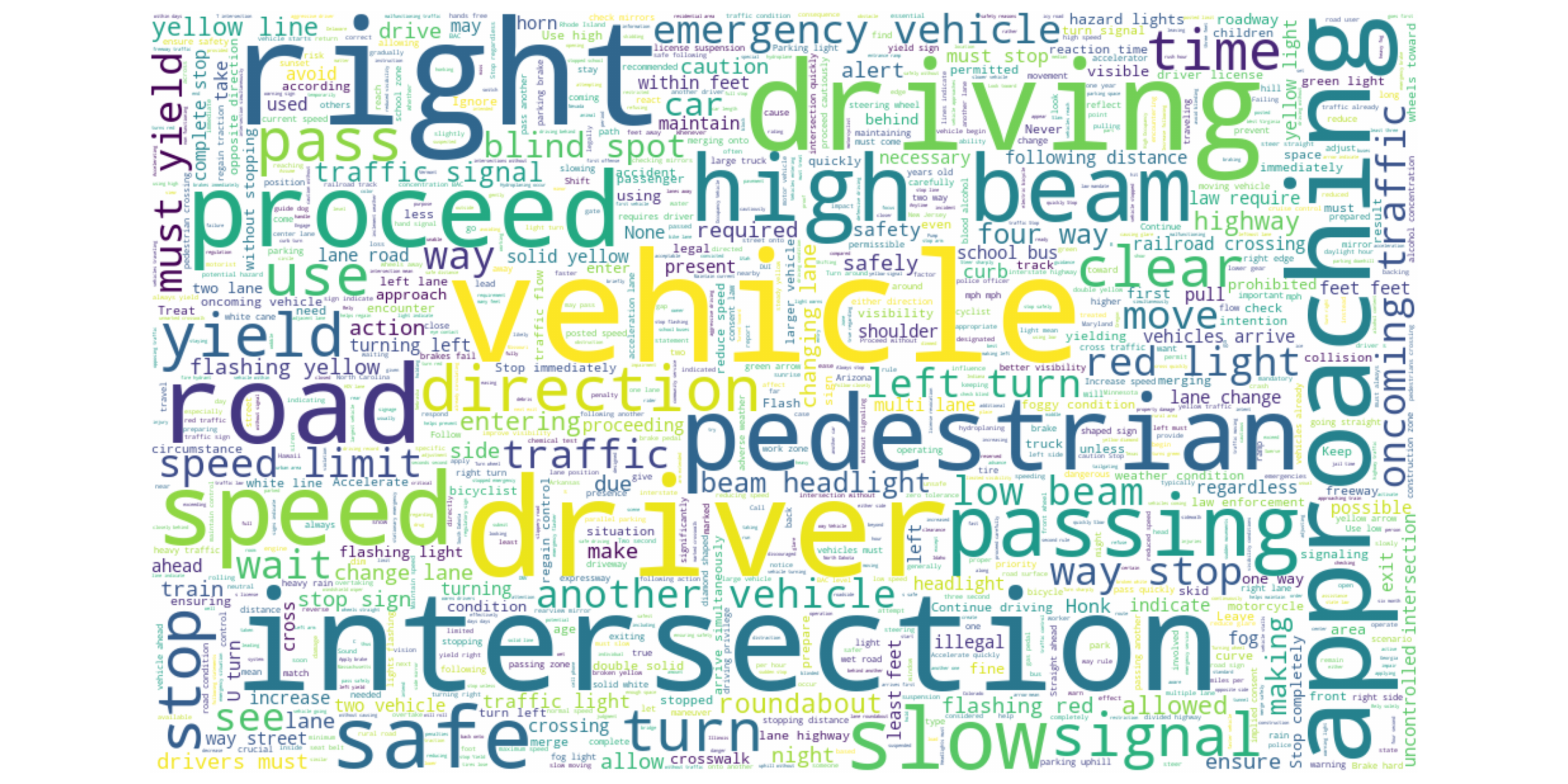}
    \vspace{-20pt}
    \caption{\textbf{Word Cloud of Questions in DriveQA.} The figure statistically summarizes the language terms in the introduced DriveQA benchmark. }
    \label{fig:textdata_word_cloud}
    \vspace{-4mm}
\end{figure}

\boldparagraph{Text-based QA Dataset---DriveQA-T} Our DriveQA-T dataset contains a total of 26K QA pairs covering different general driving topics, including traffic lights, traffic signs, parking, regulation, and symbols (see our supplementary for full details on the categories). 
Each QA pair contains an explanation for the correct answer, which can be used to evaluate the reasoning capabilities of LLMs. To curate the QA pairs, we first gathered 51 official driver's handbooks from all 50 US states plus DC. Although our data set is US-centric, it can inform the construction of additional international datasets in the future. We build DriveQA-T in three steps. First, we generate questions automatically by prompting GPT-4o~\cite{gpt4o,zhou2024embodied} with the driver's handbooks as context, and then conduct manual quality verification based on the driver’s handbooks. Quality checks were performed in rounds, where each verifier went through questions, and then ambiguous or inconsistent cases were discarded. Additional details about this process can be found in the supplementary. We note that humans, once trained, can obtain 100\% on our benchmark. We categorized the text data into 19 classes, grouped into five main categories, as shown in Fig.~\ref{fig:textdata_pie_chart}. A summarized description of the dataset is depicted in Fig.~\ref{fig:textdata_word_cloud}, showing a focus on traffic participants and intersections (\eg, right-of-way, yielding behaviors).

\boldparagraph{Multimodal Extension With DriveQA-V} Driver knowledge tests~\cite{drivertests,drivertests2} are primarily text-based, \eg, with a full description of objects and spatial layout information in text. However, certain questions particularly related to  \textit{traffic signs} and \textit{right-of-way}, test understanding through graphical illustrations accompanying text information. DriveQA-V focuses on these two types of questions. To ensure comprehensive coverage through procedural variations (\eg, camera perspectives, time of day, weather, distance), images are collected with the open-source Unreal Engine-based CARLA simulator~\cite{Dosovitskiy17CARLA}. However, since CARLA was not originally designed with extensive traffic rule knowledge, \eg, traffic signs, we augment the simulation with additional 3D assets and automatic traffic rule scripts. Due to procedural and synthetic generation, in addition to aligned text-image VQA pairs, we are able to collect full state information, such as camera perspective, distance from ego-vehicle, and sign type. 
Specifically, we insert 220 US-based traffic sign models into the map, simulator, and spawn an ego vehicle to collect sensor readings. 
For right-of-way questions, we identify intersections in the CARLA maps and randomly spawn vehicles on each side of the intersection. Each vehicle varies in color to facilitate identification in the questions.

\section{Method}
\label{sec:method}
In this section, we describe our approach to evaluating models on our proposed dataset. The methodology includes question-type classification, model evaluation using Chain of Thought (CoT)~\cite{wei2022chain} reasoning, Retrieval-Augmented Generation (RAG)~\cite{lewis2020retrieval} techniques, and model fine-tuning on the benchmark.

\boldparagraph{Question-Type Classification}
To precisely assess model performance across specific traffic rule categories, we divide the DriveQA-T dataset questions into types. This enables us to assess performance on specific traffic rule categories, thereby providing a nuanced understanding of how well they generalize across various traffic contexts.
Specifically, we apply hierarchical clustering~\cite{nielsen2016hierarchical} to organize questions into semantically coherent groups, ensuring that similar questions are grouped together based on their thematic content. We begin by generating embeddings for each question using BERT~\cite{kenton2019bert}, which effectively captures the semantic nuances of each question and represents them in a high-dimensional embedding space. By applying hierarchical clustering to these embeddings, we identify clusters that correspond to distinct traffic rule topics, such as traffic signals, speed limits, parking regulations, etc. To interpret and label each cluster, we use KeyBERT~\cite{grootendorst2020keybert} to extract semantic keywords for each group, combined with sample questions from each cluster, finally we assign descriptive types to the clusters. In DriveQA-V, we assign types manually (see supplementary for more details). 

\boldparagraph{Fine-Tuning} Off-the-Shelf models were trained on open web data, thus having potential access to driver handbooks and tests. To further investigate the role of training data for our task, we also fine-tune models on our dataset. We find this to enhance, but not fully address, models' ability to handle the specific complexities of traffic scenarios. We employ LoRA~\cite{hu2021lora}, which reduces the number of trainable parameters by introducing low-rank updates to the weight matrices in transformer layers, allowing efficient fine-tuning without requiring extensive computational resources. 

\boldparagraph{CoT and RAG}
We employ CoT reasoning and RAG-based context in our evaluation. CoT reasoning guides the LLMs and MLLMs through each reasoning step in a logical progression, which allows us to test their capacity for logical consistency, especially in multi-vehicle or rule-based scenarios. We also evaluate the produced reasoning, \eg, to ensure correct answers are selected for the correct reasons. For RAG, we construct a retrieval corpus derived from the official driver’s handbooks of all 50 U.S. states and DC. This corpus serves as a reliable, contextually relevant reference to provide the models with related context when answering questions. By retrieving it for each question, RAG-based context grounds the model’s responses in actual regulations, aiming to enhance both the accuracy and contextual relevance of answers.

\vspace{-5pt}
\section{Experiments}
\label{sec:exp}

\subsection{Setup}
We evaluate our dataset on various MLLMs. 
For each model type, we consider both open-source and closed-source variants, applying CoT and RAG strategies to structure the input prompts. Our evaluation is based not only on testing the original capabilities of each off-the-shelf model but also on a comprehensive analysis of their performance after fine-tuning the open-source checkpoint on our dataset. 

\boldparagraph{Prompt Structure}
We designed four prompt structures to explore model performance under varying levels of reasoning and contextual support. Beginning with a basic prompt, we tested standard question-answering without additional guidance. Building on this, we introduced a CoT prompt to encourage step-by-step reasoning, aiming to enhance answer consistency in complex scenarios. To further improve contextual relevance, we combined CoT with RAG-based context by retrieving pertinent information from drivers’ handbooks, thereby grounding the responses in real-world regulations. Finally, we assessed the impact of RAG-based context alone, where we provided retrieved contextual information without step-by-step reasoning. These four prompts allowed us to examine the models' capabilities in integrating both reasoning and factual support effectively.

\boldparagraph{Metrics}
To comprehensively evaluate our model's performance on both the DriveQA-T and DriveQA-V datasets, we use accuracy as the primary metric, reflecting the model's ability to correctly answer a wide range of driving-related questions across textual and visual domains. 
For the DriveQA-T dataset, we place an additional emphasis on reasoning capability, as each question includes an accompanying explanation. To measure the relevance of the model’s reasoning, we employ BLEU-4 \cite{papineni2002bleu} and ROUGE-L \cite{lin2004rouge}, providing insights into the model’s ability to generate responses that are not only accurate but also demonstrate high-quality reasoning aligned with expected standards.

\subsection{Results}
\vspace{-5pt}

\boldparagraph{Performance of LLMs on DriveQA-T}
Table~\ref{tab:results_text} presents the performance of various models on our DriveQA-T dataset. Phi-3.5-mini and Gemma-2 (9B) generally perform better across most categories than other models, demonstrating their ability to comprehend driving rules. Observably, models with CoT reasoning and RAG-based context tend to achieve higher accuracy, suggesting that these enhancements contribute to performance improvements. This trend highlights the importance of advanced reasoning and contextual retrieval for complex, rule-based tasks. While certain models show promising results in accurately interpreting and following traffic regulations, consistent performance across diverse driving-related categories may still require further refinement. 

\begin{table}[t!]
\centering
\caption{\textbf{Challenging Categories on DriveQA-T}.  We show the results of most difficult 3 types: Limits: Speed and Distance Limits, Parking: Parking and Wheel Positioning, Intersection: Right-of-Way and Lane Selection. The Average is the summary based on all 19 types of questions.  We denote with \textbf{green the top method}, and \textbf{light green second best}.}
\vspace{-5pt}
\begin{adjustbox}{max width=\linewidth}
\begin{tabular}{l c | c c c | c c c | c }
\toprule
\textbf{Models} & 
\textbf{Size} & 
\textbf{CoT} & 
\textbf{RAG} & 
\textbf{Finetune} & 
\textbf{Limits} &
\textbf{Parking} &
\textbf{Intersection} & 
\textbf{Average} 
\\
 \midrule
 \multirow{4}{*}{Gemma-2~\cite{team2024gemma}}
   & \multirow{4}{*}{2B} &  &  &  & 42.15 & 35.64 & 27.88 & 44.15
  \\
   &  & \checkmark &  &  &  42.98 & 42.57 & 34.51 & 52.77
  \\
   &  & \checkmark & \checkmark &  &  58.68 & 47.52 & 55.75 & 56.62
  \\
   &  & \checkmark & \checkmark & \checkmark & 62.40 & 61.39 & 85.84 & 72.01
  \\
\cmidrule(l){1-9}
\multirow{4}{*}{Gemma-2~\cite{team2024gemma}}
   & \multirow{4}{*}{9B} &  &  &  &  57.85 & 54.46  & 58.41 & 71.00
  \\
   &  & \checkmark &  &  &  59.50 & 58.42 & 62.83 & 72.20
  \\
   &  & \checkmark & \checkmark &  & 64.88 & 68.32 & 77.88 & 76.91
  \\
   &  & \checkmark & \checkmark & \checkmark & 72.31 &  \cellcolor{g0}88.12 & 91.15 & 87.28
  \\
\cmidrule(l){1-9}
\multirow{4}{*}{Llama-3.1~\cite{dubey2024llama}}
   & \multirow{4}{*}{8B} &  &  &  &   53.72 & 37.62 & 48.23 & 55.89
  \\
   &  & \checkmark &  &  & 55.37 & 38.61 & 65.93 & 56.22
  \\
   &  & \checkmark & \checkmark & &  55.37 & 46.53 & 68.58 & 60.79
  \\
   &  & \checkmark & \checkmark &  \checkmark & \cellcolor{g0}72.73 & 86.14 &\cellcolor{g0}91.59 & \cellcolor{g0}87.62
  \\
\cmidrule(l){1-9}
\multirow{4}{*}{Llama-3.2~\cite{dubey2024llama}}
   & \multirow{4}{*}{3B} &  &  & & 36.78 & 35.64 & 42.92 & 50.93
  \\
   &  & \checkmark &  & & 48.35 & 26.73 & 49.56 & 48.92
  \\
   &  & \checkmark & \checkmark &  &  61.16 & 53.47 & 61.50 & 64.19
   \\
   &  & \checkmark & \checkmark & \checkmark & 69.42 & 75.25 & 85.84 & 82.82
  \\
\cmidrule(l){1-9}
\multirow{4}{*}{Phi-3.5-mini~\cite{abdin2024phi}}
   & \multirow{4}{*}{3.8B} &  &  &  &  49.17 & 48.51 & 79.65 & 69.79
  \\
   &  & \checkmark &  &  & 55.79 & 45.54  & 79.65 & 71.14
  \\ 
   &  & \checkmark & \checkmark &  &  63.22 & 62.38 & 84.96 & 77.30
   \\
   &  & \checkmark & \checkmark & \checkmark & 66.94 & 65.35 & 87.17 & 81.08
  \\
\cmidrule(l){1-9}
GPT-4o~\cite{gpt4o} & - & \checkmark & \checkmark &  & \cellcolor{g1}76.72 & \cellcolor{g1}93.75 &\cellcolor{g1}97.27 & \cellcolor{g1}91.96
   \\
   
\bottomrule
\end{tabular}
\end{adjustbox}
\vspace{-5pt}
\label{tab:results_text}
\end{table}

\begin{table}[t!]
\caption{\textbf{Performance of CoT Reasoning on DriveQA-T.} The evaluation includes both off-the-shelf and fine-tuned models under two settings of with and without RAG. }
\vspace{-5pt}
\label{tab:result_00_baselines}
\resizebox{0.48\textwidth}{!}{%
\begin{tabular}{ll|cc|cc}
\toprule
  \multirow{2}{*}{\textbf{Models}} &
  \multirow{2}{*}{\textbf{Size}} &
  \multicolumn{2}{c}{\textbf{BLEU-4}} & 
  \multicolumn{2}{c}{\textbf{ROUGE-L}} \\
  \cmidrule(lr){3-6} 
  & & \textbf{w/o RAG} & \textbf{w/ RAG} & \textbf{w/o RAG} & \textbf{w/ RAG} \\
\midrule
\multicolumn{6}{c}{\textit{Off-The-Shelf Models}} \\
\midrule
Gemma-2~\cite{team2024gemma} & 2B  & 0.1098 & 0.1704 & 0.2920 & 0.3387 \\
Gemma-2~\cite{team2024gemma} & 9B  & \cellcolor{g0}0.3234 & 0.3116 & \cellcolor{g0}0.4295 & \cellcolor{g0}0.4276 \\
Llama-3.1~\cite{dubey2024llama} & 8B  & 0.2573 & 0.2619 & 0.3270 & 0.3317 \\
Llama-3.2~\cite{dubey2024llama} & 3B & 0.2258 & \cellcolor{g0}0.3140 & 0.3348 & 0.4024 \\
Phi-3.5-mini~\cite{abdin2024phi} & 3.8B & 0.2437 & 0.2574 & 0.3616 & 0.3996 \\
GPT-4o~\cite{gpt4o} & -  & \cellcolor{g1}0.3905 & \cellcolor{g1}0.3989 & \cellcolor{g1}0.5354 & \cellcolor{g1}0.5393 \\

\midrule
\multicolumn{6}{c}{\textit{Finetuned Models}} \\
\midrule
Gemma-2~\cite{team2024gemma} & 2B & \cellcolor{g0}0.3623 & 0.2934 & \cellcolor{g0}0.5058 & 0.4458 \\
Gemma-2~\cite{team2024gemma} & 9B & \cellcolor{g1}0.4112 & \cellcolor{g1}0.4105 & \cellcolor{g1}0.5420 & \cellcolor{g1}0.5528 \\
Llama-3.1~\cite{dubey2024llama} & 8B & 0.3042 & \cellcolor{g0}0.2946 & 0.4749 & \cellcolor{g0}0.4750 \\
Llama-3.2~\cite{dubey2024llama} & 3B & 0.2131 & 0.1916 & 0.3853 & 0.3570 \\
Phi-3.5-mini~\cite{abdin2024phi} & 3.8B & 0.2362 & 0.1891 & 0.4073 & 0.3476 \\

\bottomrule
\end{tabular}%
}
\vspace{-5pt}
\label{tab:results_cot}
\end{table}

\begin{table*}[t!]
\centering
\caption{\textbf{Summarized Results on DriveQA-V.} We show model performance (accuracy \%) for VQA. The dataset is divided into two main categories: intersections and signs (categorized into camera perspective and type). 
}
\vspace{-5pt}
\begin{adjustbox}{max width=\textwidth}
\begin{tabular}{l l | c c c c | c c c c | c}
\toprule
\multirow{2}{*}{\textbf{Models}} & 
\multirow{2}{*}{\textbf{Size}} & 
\multicolumn{4}{c}{\textbf{DriveQA-V (Inters.)}} & 
\multicolumn{4}{c}{\textbf{DriveQA-V (Signs)}} & 
\multirow{2}{*}{\textbf{Average}} \\ 
\cmidrule(lr){3-10}
& & \textbf{T-Front} & \textbf{T-Top} & \textbf{Cross-Front} & \textbf{Cross-Top} & \textbf{Regulatory} & \textbf{Warning} & \textbf{Guide} & \textbf{Temporary Control} \\

\midrule
\multicolumn{11}{c}{\textit{Off-The-Shelf Models}} \\
\midrule
Mini-InternVL~\cite{gao2024mini} & 2B & \cellcolor{g0}27.83 & \cellcolor{g0}24.83 & 26.00 & 25.65 & \cellcolor{g0}64.06 & \cellcolor{g0}55.34 & \cellcolor{g0}65.82 & \cellcolor{g0}45.04 & \cellcolor{g0}41.82 \\
LLaVA-1.5~\cite{liu2024improved} & 7B & 23.30 & 23.10 & 24.96 & 23.24 & 23.51 & 26.61 & 22.31 & 21.10 & 23.52 \\
LLaVA-1.6-mistral~\cite{liu2024llavanext} & 7B & 18.77 & 19.66 & \cellcolor{g0}30.99 & \cellcolor{g0}30.47 & 42.58 & 43.01 & 52.75 & 37.50 & 34.47 \\
VILA-1.5~\cite{lin2024vila} & 8B & 15.53 & 16.86 & 15.69 & 20.35 & 25.32 & 23.33 & 27.78 & 21.46 & 20.79 \\
GPT-4o~\cite{gpt4o} & - & \cellcolor{g1}55.09 & \cellcolor{g1}60.36 & \cellcolor{g1}50.52 & \cellcolor{g1}59.14 & \cellcolor{g1}93.75 & \cellcolor{g1}94.02 & \cellcolor{g1}95.11 & \cellcolor{g1}94.35 & \cellcolor{g1}75.29 \\
\midrule
\multicolumn{11}{c}{\textit{Finetuned Models}} \\
\midrule
Mini-InternVL~\cite{gao2024mini} & 2B & \cellcolor{g1}86.73 & \cellcolor{g0}82.07 & \cellcolor{g0}74.33 & \cellcolor{g1}76.01 & \cellcolor{g1}93.79 & \cellcolor{g1}92.19 & \cellcolor{g0}91.08 & \cellcolor{g1}96.51 & \cellcolor{g1}86.59 \\
LLAVA-1.5~\cite{liu2024improved} & 7B & 64.18 & 70.57 & 54.77 & 56.52 & 72.22 & 73.00 & 76.82 & 89.27 & 69.67 \\
LLaVA-1.6-mistral~\cite{liu2024llavanext} & 7B & \cellcolor{g0}86.08 & \cellcolor{g1}85.52 & \cellcolor{g1}74.38 & \cellcolor{g0}74.53 & 82.05 & \cellcolor{g0}84.10 & 88.11 & 94.49 & \cellcolor{g0}83.66 \\
VILA-1.5~\cite{lin2024vila} & 8B & 47.67 & 52.27 & 55.60 & 57.26 & \cellcolor{g0}87.10 & 83.14 & \cellcolor{g1}91.46 & \cellcolor{g0}95.33 & 71.23\\

\bottomrule
\end{tabular}
\end{adjustbox}
\vspace{-5pt}
\label{tab:results_visual}
\end{table*}

\begin{table*}[t!]
    \centering
    \caption{\textbf{10 Most Difficult Sign Types in DriveQA-V.}
    We calculate the lowest accuracy over all the models' performance based on different sign types. Most challenging cases belong to regulatory and warning signs. } 
    \vspace{-2mm}
    \begin{adjustbox}{max width=\textwidth}
    \begin{tabular}{l c | c c c c c c c c c c}
        \toprule
        \multirow{2}{*}{\textbf{Model}} & \multirow{2}{*}{\textbf{Size}} & \multirow{2}{*}{\textbf{Playground}} & \textbf{Trauma} & \textbf{Golf} & \textbf{Ground} & \textbf{No} & \textbf{No} & \textbf{Push} & \textbf{Weekday} & \textbf{Fire} & \textbf{Tractor} \\
        & & & \textbf{Center} & \textbf{Carts} & \textbf{Clearance} & \textbf{Stopping} & \textbf{Parking} & \textbf{Button} & \textbf{Only} & \textbf{Truck} & \textbf{Crossing}\\
        \midrule
        \multicolumn{11}{c}{\textit{Off-The-Shelf Models}} \\
        \midrule
        Mini-InternVL~\cite{gao2024mini} & 2B & 0.00 & \cellcolor{g1}27.78 & \cellcolor{g0}14.81 & \cellcolor{g0}15.00 & \cellcolor{g1}42.10 & \cellcolor{g1}48.14 & 0.00 & 8.70 & 0.00 & 20.00 \\
        LLaVA-1.5~\cite{liu2024improved} & 7B & \cellcolor{g1}5.26 & 2.38 & 5.43 & 11.76 & \cellcolor{g0}16.30 & 10.42 & \cellcolor{g0}12.50 & \cellcolor{g1}20.83 & \cellcolor{g1}20.45 & 21.59 \\
        LLaVA-1.6-mistral~\cite{liu2024llavanext} & 7B & 0.00 & 16.67 & \cellcolor{g1}25.93 & \cellcolor{g1}25.00 & 5.00 & \cellcolor{g0}22.22 & \cellcolor{g1}48.00 & 17.39 & 0.00 & \cellcolor{g0}24.00 \\
        VILA-1.5~\cite{lin2024vila} & 8B & \cellcolor{g0} 1.32 & \cellcolor{g0}2.38 & 0.00 & 8.82 & 8.70 & 5.21 & 2.78 & \cellcolor{g0}19.79 & \cellcolor{g0}3.41 & \cellcolor{g1}31.82 \\
        \midrule
        \multicolumn{11}{c}{\textit{Finetuned Models}} \\
        \midrule
        Mini-InternVL~\cite{gao2024mini} & 2B & \cellcolor{g1}88.46 & \cellcolor{g1}94.44 & \cellcolor{g1}88.88 & \cellcolor{g1}95.00 & \cellcolor{g1}100.00 & \cellcolor{g1}90.91 & \cellcolor{g1}96.00 & \cellcolor{g1}100.00 & \cellcolor{g1}92.86 & \cellcolor{g0}80.00 \\
        LLaVA-1.5~\cite{liu2024improved} & 7B & 73.68 & \cellcolor{g0}85.71 & 61.96 & 64.71 & 65.22 & 61.46 & 75.00 & 37.50 & 59.09 & 57.95 \\
        LLaVA-1.6-mistral~\cite{liu2024llavanext} & 7B & \cellcolor{g0}80.77 & 83.33 & 74.07 & \cellcolor{g0}85.00 & \cellcolor{g0}85.19 & 77.27 & \cellcolor{g0}92.00 & \cellcolor{g1}100.00 & \cellcolor{g1}92.86 & \cellcolor{g1}92.00 \\
        VILA-1.5~\cite{lin2024vila} & 8B & 68.42 & 59.52 & \cellcolor{g0}83.70 & 66.18 & 65.22 & \cellcolor{g0}79.17 & 66.67 & \cellcolor{g0}80.21 & \cellcolor{g0}80.68 & 52.27 \\
        \bottomrule
    \end{tabular}%
    \end{adjustbox}
\label{tab:results_sign_types}
\vspace{-5pt}
\end{table*}

As shown in Table~\ref{tab:results_text}, all models exhibit a significant improvement in overall accuracy after fine-tuning. However, they still struggle with numerical questions, such as those in the ``\texttt{Limits}'' and ``\texttt{Alcohol}'' categories. This difficulty suggests that models may lack the precise numerical reasoning capabilities needed to respond accurately to questions involving specific values or quantitative thresholds, which are critical in understanding speed limits, alcohol levels, and other regulatory metrics. Furthermore, for certain decision-making-focused categories, including ``\texttt{Passing}'', ``\texttt{Signs}'' and ``\texttt{Turning}'', most models achieve only slightly above accuracy of 80\%. These categories are crucial for safe driving in practical conditions, highlighting the models' continuous shortcomings in handling nuanced, context-dependent traffic rules despite fine-tuning improvements.

\boldparagraph{CoT Reasoning of LLMs on DriveQA-T}
Table~\ref{tab:results_cot} shows the evaluation results of CoT reasoning on the DriveQA-T dataset. Most models show improvements when using RAG-based context. Specifically, GPT-4o achieves the highest BLEU-4 and ROUGE-L scores among the off-the-shelf models, reaching a BLEU-4 score of 0.3989 and a ROUGE-L score of 0.5393 with RAG-based context. After fine-tuning, Gemma-2 (9B) surpasses GPT-4o in both BLEU-4 and ROUGE-L scores, demonstrating the effectiveness of fine-tuning in adapting the model specifically to traffic rules and enabling it to provide more accurate, context-specific explanations. However, these scores still fall short of what would be considered high-quality for generating fully robust and exhaustive explanations, indicating that the models are not yet capable of consistently producing complete and nuanced responses.
Furthermore, the lower scores of Llama-3.2 and Phi-3.5-mini after fine-tuning suggest potential issues. One possible reason for this decline is overfitting the fine-tuning dataset, which may cause the models to become too specialized and lose some of their generalization capabilities. This overfitting can result in explanations that are overly tailored to specific training examples, reducing the models' ability to produce flexible, broadly applicable responses. Additionally, fine-tuning may interfere with the effectiveness of RAG-based retrieval, leading to less relevant contextual information and, consequently, lower alignment with ground-truth explanations. These factors highlight the challenges of balancing specificity and generalization in fine-tuning for complex, rule-based tasks.

\boldparagraph{Performance of MLLMs on DriveQA-V}
Table~\ref{tab:results_visual} presents the accuracy of MLLM models on DriveQA-V, which assesses model performance across intersection types and traffic sign categories. The dataset divides intersections into 4 different categories based on the intersection types and camera perspective, and 4 different categories of signs based on most states' driver handbooks. 
Among the off-the-shelf models, GPT-4o achieves the highest accuracy in all intersection and sign categories, with a particularly strong performance in the sign types (around 94\%). This suggests that GPT-4o possesses a deep understanding of signs. However, for intersection-based categories, the performance remains relatively low, with the highest off-the-shelf accuracy of 60.36\% in the ``\texttt{T-Top}'' category. Most models except GPT-4o perform below random guess level (25\%) in several categories due to bias~\cite{tong2024eyes}. This indicates that off-the-shelf models struggle to fully understand and apply traffic rules in intersection scenarios, which often require more complex visual-spatial reasoning.
Additionally, Fine-tuning significantly enhances model performance across all categories. All models achieve notable improvements after fine-tuning, which demonstrates that fine-tuning effectively adapts MLLMs to handle the visual-spatial and contextual nuances for the accurate understanding of both right-of-way rules and traffic signs.

Despite these gains, there remain limitations. Both LLaVA-1.5 and VILA-1.5, even after fine-tuning, achieve only moderate accuracy in intersection categories, with particularly lower performance on first-person perspective images. This suggests that the models still struggle with complex, multi-vehicle intersection scenarios, where perspective and spatial relationships are critical. 
For the traffic signs recognition task, We can observe the best training performance in the Guide Signs and Temporary Traffic Control categories. This is because guide signs typically feature simpler images with blue backgrounds, while temporary traffic control signs have distinct orange backgrounds and normally larger sign sizes, making them easier for the model to learn and generalize. However, many critical traffic signs fall under the Regulatory and Warning categories, including speed limit, no entry, etc. As shown in Table~\ref{tab:results_sign_types}, among the ten worst-performing sign types, only ``Trauma Center'' belongs to the Guide Signs category, with the most challenging signs coming from the Regulatory and Warning categories. This highlights significant room for improvement in the current visual model. While fine-tuned models perform well on ``\texttt{Guide}'' and ``\texttt{Temporary Control}'' signs, their performance does not consistently exceed 90\%.
Based on both Table~\ref{tab:results_text} and Table~\ref{tab:results_visual} and shown in categories' accuracy, the zero-shot performance on DriveQA-V is much lower than on DriveQA-T. This indicates that current MLLMs' fine-grained perception and visual reasoning capabilities are nascent, exhibiting systematic shortcomings due to CLIP's failures.

\boldparagraph{Role of Difficulty and Distractors}
To further increase the evaluation difficulty, we adopt a negative sampling strategy to construct more challenging distractors. Specifically, for DriveQA-T, we construct a difficult question set containing 1249 questions. For DriveQA-V (Signs), we leverage metadata, \ie, the ground-truth traffic sign artifact categories to ensure that distractors belong to the same category as the correct answer. For numeric signs, all candidates are constrained to numerical values to further increase ambiguity. Evaluation results on GPT-4o and a representative open-source baseline are summarized in Table~\ref{tab:results_negative_sampling}.

\begin{table}[t!]
\centering
\caption{{\textbf{Role of Difficult Questions and Distractors.} The accuracy degradation on a hard subset of DriveQA-T and on a challenging set of DriveQA-V with negative sampling shows the limitations of current models, including GPT-4o, in accurately understanding complex traffic rules and signs.}}
\vspace{-5pt}
\begin{adjustbox}{max width=1.0\linewidth}
\renewcommand{\arraystretch}{0.6}
\small
\begin{tabular}{c l c c c c}
\toprule
\multirow{2}{*}{\textbf{Test Set}} & 
\multirow{2}{*}{\textbf{Models}} & 
\multirow{2}{*}{\textbf{Size}} & 
\multicolumn{2}{c}{\textbf{Neg. Sampling}} & \multirow{2}{*}{\textbf{Degradation}} \\ 
\cmidrule{4-5}
 & & & \textbf{Before} & \textbf{After}  \\
\midrule
\multirow{2}{*}{DriveQA-T} & Llama-3.1~\cite{dubey2024llama} & 8B & {55.89} & 39.87 & 28.66\% \\
 & GPT-4o~\cite{gpt4o} & - & {91.96} & 78.91 & 14.19\% \\
\midrule
 \multirow{2}{*}{DriveQA-V (Signs)}& LLaVA-1.5~\cite{liu2024improved} & 13B & {11.92} & 9.82 & 17.62\% \\
 & GPT-4o~\cite{gpt4o} & - & {94.10} & 79.40 & 15.62\% \\
\bottomrule
\end{tabular}
\end{adjustbox}
\label{tab:results_negative_sampling}
\vspace{-2mm}
\end{table}

\begin{table}[t!]
\centering
\caption{\textbf{Sim-to-Real Generalization.} We pre-train on synthetic DriveQA (DQA) and evaluate on real-world Mapillary images. The Mapillary dataset comprises challenging scenarios with various traffic sign placements, occlusion, and illumination. }
\vspace{-5pt}
\begin{adjustbox}{width=0.48\textwidth}
\begin{tabular}{c | l  c | c c}
\toprule
\multirow{2}{*}{\textbf{Test Set}} & 
\multirow{2}{*}{\textbf{Models}} & 
\multirow{2}{*}{\textbf{Size}} & 
\multicolumn{2}{c}{\textbf{Accuracy}} \\ 
\cmidrule{4-5}
 & & & \textbf{Off-The-Shelf} & \textbf{DQA-Finetuned} \\
\midrule
 \multirow{5}{*}{Real-World Mapillary~\cite{neuhold2017mapillary}}
 & Mini-InternVL~\cite{gao2024mini} & 2B & \cellcolor{g0}57.25 & \cellcolor{g1}68.61 \\
 & LLaVA-1.5~\cite{liu2024improved} & 7B & 40.68 & 52.34 \\
 & LLaVA-1.6-mistral~\cite{liu2024llavanext} & 7B & 53.18 & 57.71 \\
 & VILA-1.5~\cite{lin2024vila} & 8B & 34.38 & \cellcolor{g0}60.86 \\
 & GPT-4o~\cite{gpt4o} & - & \cellcolor{g1}84.73 & - \\
\bottomrule
\end{tabular}
\end{adjustbox}
\vspace{-2mm}
\label{tab:real_world_mapillary}
\end{table}

\begin{table}[t!]
\centering
\caption{\textbf{End-to-End Trajectory Planning Results on nuScenes.} We compute the L2 error at different prediction horizons (1s, 2s, and 3s). Lower L2 error shows our DriveQA (DQA) dataset can transfer from simulation to real-world driving tasks.}
\vspace{-5pt}
\begin{adjustbox}{width=\linewidth}
\begin{tabular}{l | c | c c c c}
\toprule
\multirow{2}{*}{\textbf{Model}} & 
\multirow{2}{*}{\textbf{Pretrained on DQA}} & 
\multicolumn{4}{c}{\textbf{L2($m$)$\downarrow$}} \\
& & 1$s$ & 2$s$ & 3$s$ & 
Avg.
\\
 \midrule
 
 LLaVA-1.6-mistral~\cite{liu2024llavanext} (OpenEMMA~\cite{xing2025openemma})
   &  & \cellcolor{g0} 1.49 & 3.38 & 4.09 & 2.98 \\
LLaVA-1.6-mistral~\cite{liu2024llavanext}
  & \checkmark  & \cellcolor{g1} 1.30 & 3.46 & \cellcolor{g0} 3.98 & \cellcolor{g0} 2.91
  \\
\midrule
{\multirow{2}{*}{InternVL-2.5-8B} }
  &     &  1.66 & \cellcolor{g0}3.36 & 4.15 & 3.06 \\
  & \checkmark    &  \cellcolor{g1} 1.30 & \cellcolor{g1} 3.08 & \cellcolor{g1}3.73 & \cellcolor{g1}2.71 
  \\
\bottomrule
\end{tabular}
\end{adjustbox}
\label{tab:results_openemma}
\vspace{-2mm}
\end{table}

\begin{table}[t!]
\centering
\caption{{\textbf{Evaluation on BDD-OIA Dataset~\cite{xu2020explainable}.} We report mean F1 score (mF1) and overall F1 score (F1\textsubscript{all}) for both action and explanation tasks. The results show that fine-tuning on DriveQA improves performance on both tasks.}}
\vspace{-5pt}
\begin{adjustbox}{width=\linewidth}
\begin{tabular}{l | c  c | c c c c }
\toprule
\multirow{2}{*}{\textbf{Model}} & 
\multicolumn{2}{c|}{\textbf{Finetune}} &
\multicolumn{2}{c}{\textbf{Action}} & 
\multicolumn{2}{c}{\textbf{Explanation}} \\
\cmidrule{2-7}
& DQA & BDD-OIA & mF1 $\uparrow$ & F1\textsubscript{all} $\uparrow$ & mF1 $\uparrow$ & F1\textsubscript{all} $\uparrow$
\\
 \midrule

\multirow{4}{*}{InternVL-2.5-8B}
  &   &   & 0.2951 & 0.554 & 0.0624 & 0.2223 \\
  & \checkmark &  & 0.2226  & 0.4103 & 0.1549 & 0.1850 \\
  &  & \checkmark & \cellcolor{g0} 0.4911  & \cellcolor{g0} 0.7072 & \cellcolor{g0} 0.2872 & \cellcolor{g0} 0.5015 \\
  & \checkmark  & \checkmark & \cellcolor{g1}0.5285 & \cellcolor{g1}0.7334 & \cellcolor{g1}0.3102 & \cellcolor{g1}0.5448  \\
\bottomrule
\end{tabular}
\end{adjustbox}
\label{tab:results_bddoia}
\vspace{-2mm}
\end{table}

\boldparagraph{Sim-to-Real Transferability}
We evaluate our models finetuned on DriveQA on a curated dataset by us from Mapillary~\cite{neuhold2017mapillary} (1303 annotated images, including 166 sign types), as shown in Table~\ref{tab:real_world_mapillary}. Additionally, results in Table~\ref{tab:results_openemma} show the downstream trajectory planning task with OpenEMMA~\cite{xing2025openemma} on nuScenes dataset, where our task-agnostic QA model is intentionally only fine-tuned on DriveQA but tested zero-shot in waypoint prediction to measure generalization. Reduced L2 errors show the transferability of our dataset. However, nuScenes lacks diversity and is generally uneventful (\eg, minimal signage), while our benchmark exhaustively covers all traffic rules and scenarios. We therefore also make evaluations on the more diverse datasets of BDD-OIA~\cite{xu2020explainable} as shown in Table~\ref{tab:results_bddoia}. After fine-tuning on DriveQA, the models achieve better performance in cross-domain real-world driving tasks, demonstrating the effectiveness of our data in improving the understanding of traffic rules and real-world generalizability. We provide additional analysis in the supplementary.

\boldparagraph{Limitation}
While our benchmark, models, and analysis provide insights into the performance of models in understanding diverse traffic rules for autonomous driving, there are several limitations, which we plan to address in future work. First, the benchmark primarily evaluates static, structured knowledge of traffic rules. While this is aligned with standard driving knowledge tasks, there is an opportunity to leverage video-based models in the future (\eg, using our augmented CARLA simulation). Our analysis demonstrates that incorporating knowledge from text does indeed transfer to dynamic settings in nuScenes, yet vision-based reasoning remains nascent in MLLMs (or even spatial reasoning~\cite{tong2024eyes}). 
Moreover, our study highlights weaknesses in numerical reasoning and spatial awareness yet does not explore potential mitigation strategies beyond fine-tuning. The reliance on synthetic data also raises concerns about domain adaptation. Nonetheless, simulation data is crucial for scalability, as we are able to control for various variations, including occlusions and ambiguous signage, which may be rare in real-world benchmarks. 
Finally, while the dataset includes controlled variations in environmental factors like lighting and weather, it does not extensively cover edge cases such as emergency vehicle interactions (only covered in DriveQA-T) or pedestrian intent recognition.  
The models also exhibit biases towards frequently seen traffic patterns, which may result in poor generalization to geographically diverse driving environments with different road layouts and regulations.

\section{Conclusion}
\label{sec:conclusion}

In this paper, we introduce DriveQA, a novel benchmark for autonomous driving that evaluates models through text-based (DriveQA-T) and visual-text (DriveQA-V) question-answering, focusing on general traffic rules, traffic signs, and complex right-of-way scenarios. Our evaluation of state-of-the-art models reveals critical limitations: even fine-tuned models struggle with nuanced right-of-way scenarios, falling short of the reasoning needed for safe driving guidance. Our work deliberately focuses on static visual and textual inputs, \ie, to align with real-world driver knowledge tests. While video-based learning is not required to adhere to these standards, future research could explore hybrid frameworks incorporating video to address time-dependent scenarios. 
Ultimately, while humans can learn traffic rules through textual instruction and contextual practice, current models remain overly reliant on observational training data. Models thus lack the ability to internalize explicit textual knowledge and apply it effectively in decision-making. This suggests that learning traffic rules from text remains an underexplored paradigm, highlighting the need for methods that better integrate language understanding with spatial reasoning.

\boldparagraph{Acknowledgments}
We thank the National Science Foundation (award IIS-2152077) and Red Hat Collaboratory (award \#2024-01-RH02) for supporting this research.

\newpage{
    \small
    \bibliographystyle{ieeenat_fullname}
    \bibliography{main}
}


\end{document}